\def\year{2021}\relax
\documentclass[letterpaper]{article} % DO NOT CHANGE THIS
\usepackage{aaai21}  % DO NOT CHANGE THIS
\usepackage{times}  % DO NOT CHANGE THIS
\usepackage{helvet} % DO NOT CHANGE THIS
\usepackage{courier}  % DO NOT CHANGE THIS
\usepackage[hyphens]{url}  % DO NOT CHANGE THIS
\usepackage{graphicx} % DO NOT CHANGE THIS
\usepackage{graphicx}  % DO NOT CHANGE THIS
\usepackage{natbib}  % DO NOT CHANGE THIS AND DO NOT ADD ANY OPTIONS TO IT
\usepackage{caption} % DO NOT CHANGE THIS AND DO NOT ADD ANY OPTIONS TO IT
\usepackage[switch]{lineno}  %

\usepackage{multirow}
\usepackage{algorithm}
\usepackage{algorithmicx}
\usepackage{algpseudocode}
\usepackage{amsmath}
\usepackage{wrapfig}
\usepackage{graphicx}
\usepackage{booktabs}
\title{Learning with Retrospection}
\author {
        Xiang Deng,
        Zhongfei Zhang \\
}
\affiliations {
    Computer Science Department, State University of New York at Binghamton \\
    xdeng7@binghamton.edu, zhongfei@cs.binghamton.edu
}

\begin{document}
%\linenumbers  %

\maketitle

\begin{abstract}
%Deep neural networks (DNNs) have been successfully applied to various domains in artificial intelligence, including computer vision and natural language processing.
%In order to further boost the performances, the naive way is to train deeper or wider networks, which leads to an exponential increment of the computation and storage cost.
%It is appealing to further improve the DNN performance, thus benefiting the applications and extending the applications' boundaries to accuracy-crucial domains.
% such as automated driving and medical image analysis.
%LWR takes advantage of the learned instance-to-class similarities in the past epochs to guide the training in the future epochs, thus improving the performance.
Deep neural networks have been successfully deployed in various domains of artificial intelligence, including computer vision and natural language processing.
We observe that the current standard procedure for training DNNs discards all the learned information in the past epochs except the current learned weights.
An interesting question is: is this discarded information indeed useless?
We argue that the discarded information can benefit the subsequent training.
In this paper, we propose learning with retrospection (LWR) which makes use of the learned information in the past epochs to guide the subsequent training.
LWR is a simple yet effective training framework to improve accuracies, calibration, and robustness of DNNs without introducing any additional network parameters or inference cost, only with a negligible training overhead.
Extensive experiments on several benchmark datasets demonstrate the superiority of LWR for training DNNs.
\end{abstract}

\section{Introduction}
Deep neural networks (DNNs) have been successfully applied to a wide range of applications in artificial intelligence, such as automated vehicle control \cite{levinson2011towards}, biometric recognition \cite{lawrence1997face}, and medical diagnosis \cite{miotto2016deep}.
For these applications, classification is a fundamental and important task.
It is appealing to further improve the classification performance, thus benefiting these applications and extending the application horizon to more accuracy-critical or safety-critical domains.
%However, DNNs with large amounts of parameters suffer from the overfitting or overconfidence issue on classification.
%Many efforts have been made on addressing this issue, thus improving the performances.
%of DNNs, 

%However, recent studies have shown that DNNs are vulnerable to label corruption 
%including computer vision and natural language processing.
%Using DNNs to do classification is a fundamental and important task as it has varieties of applications.
%Yet, in accuracy-crucial and robustness-crucial applications such as automated vehicle control, security forensics, and medical diagnosis [26], it is appealing and important to further improve the classification performance of DNNs.
%DNNs with millions of parameters may suffer from overfitting and overconfidence.
%Many efforts have been made to address this problem.
%Among these applications, classification is a fundamental and important task.
\par

\begin{figure}[t]
\setlength{\abovecaptionskip}{0.2cm}
\setlength{\belowcaptionskip}{-0.6cm}
\centering
     \includegraphics[width=0.47\textwidth]{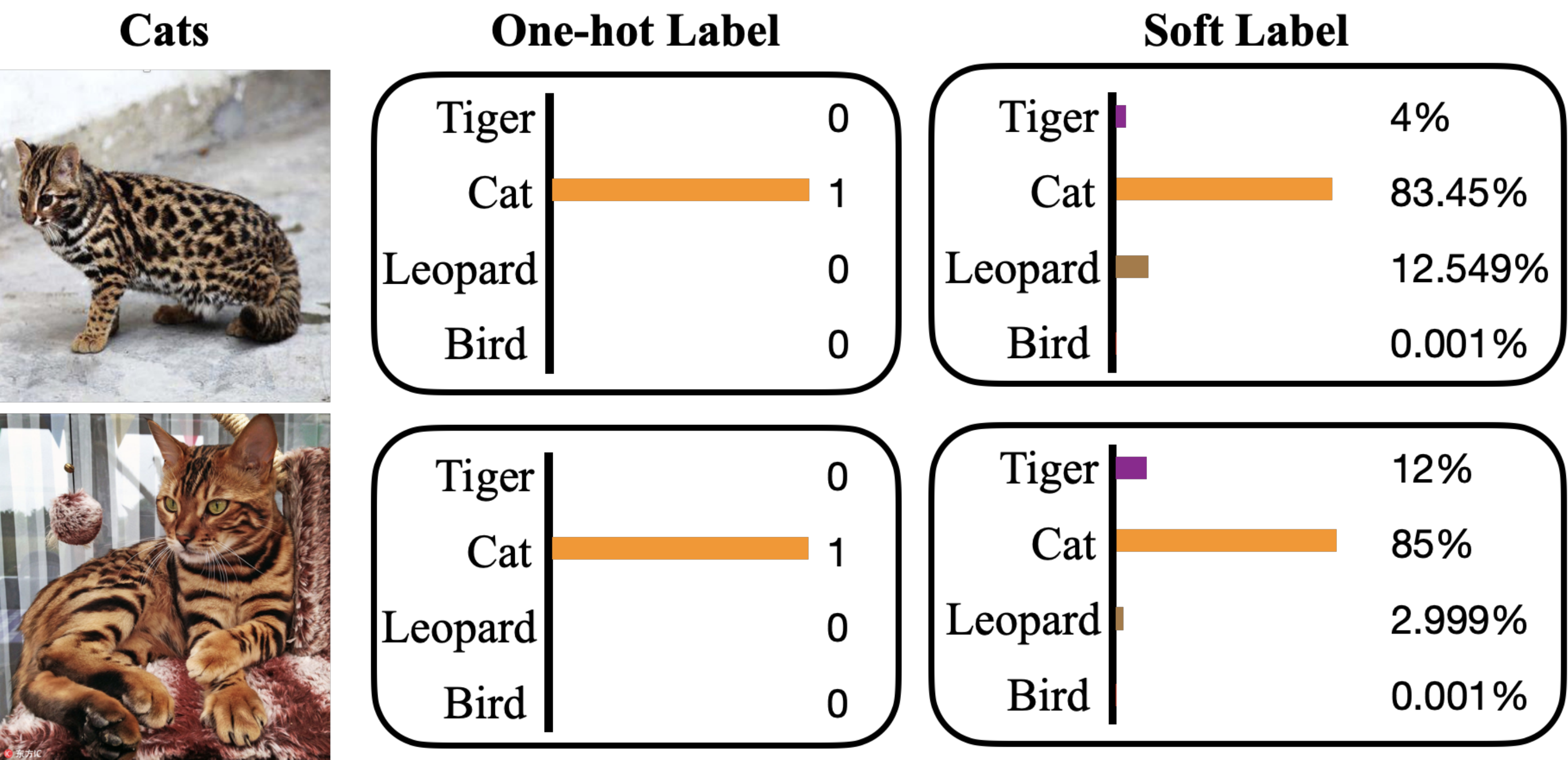}
     \caption{As seen from the coloration patterns of the two cats, the cat in the first row is more visually similar to a leopard while the cat in the second row is more visually similar to a tiger.
     Instead of assigning them the same one-hot label, it is reasonable to assign them different soft labels with different probabilities for different classes.}
     \label{f1}
\end{figure}

The standard procedure for training DNNs on classification is to fit DNN outputs to one-hot labels by using the cross-entropy loss.
In a one-hot label, the probability (confidence score) for the ground-truth class is set to 1 while the probabilities for the other classes are all 0s, which means that the labels for different classes are orthogonal.
However, it is observed that the individuals in different classes usually share visual or semantic similarities, e.g., cats may be visually similar to tigers or leopards.
This indicates that one-hot labels are not necessarily the optimal target to fit as they ignore the class similarity information.
Thus, besides maximizing the probability of the ground-truth class, allowing for probabilities of the other classes (which may be visually or semantically similar to the ground-truth) to be preserved is helpful for mitigating the risk of over-fitting or over-confidence.
%helps to alleviate the risk of over-fitting or over-confidence.
Motivated by this observation, we turn to soft labels.
Two intuitive examples about one-hot and soft labels are shown in Figure \ref{f1}.
For the two cats in Figure \ref{f1}, the one-hot labels set the probability for the cat class to 1 and those for the other classes to 0, which ignores the class similarities.
In contrast, soft labels are able to take into account instance-to-class similarities.

%In line with the above idea, 
Coinciding with the above idea, many approaches have benefited from soft labels which assign small probabilities to non-ground-truth classes.
The label smoothing regularizer (LSR) \cite{muller2019does,szegedy2016rethinking} which uses soft labels generated by interpolating one-hot labels and uniform-distribution labels has improved the performances on image classification \cite{szegedy2016rethinking,zoph2018learning,real2019regularized}, speech recognition \cite{chorowski2016towards}, and machine translation \cite{vaswani2017attention}, but LSR is unable to capture instance-to-class similarities, e.g., some cats are more visually similar to tigers while some other cats are more similar to leopards as illustrated in Figure \ref{f1}.
Knowledge distillation (KD) \cite{hinton2015distilling} takes advantage of the instance-level soft labels generated by a pretrained teacher network to train a student network, thus improving the student performance significantly.
However, the training cost of KD is several times of that of the standard training procedure due to two factors: (1) KD first needs to train a teacher network; (2) when KD trains the student network, in each iteration, each image needs to be processed twice, once by the teacher network and once by the student network.
\par

Instead of using a manually designed uniform distribution or a pretrained DNN to generate soft labels, we turn to the learned instance-to-class similarities in the past epochs.
In the standard training procedure, all the learned information in the earlier epochs except the weights is discarded and the training process progresses by using one-hot labels all the time.
Different from the standard training procedure, we propose to learn with retrospection (LWR) to make use of the learned information in the past epochs.
Specifically, LWR generates and then updates the soft labels for each image by taking advantage of the output logits in the past epochs.
%the learned instance-to-class similarities
The soft labels are then used to guide the training in the subsequent epochs, thus mitigating the overconfidence or overfitting issue.
LWR is able to improve accuracies, robustness, and calibration of DNNs without introducing any additional network parameters, without increasing inference cost, and even without needing to further tune training hyperparameters (i.e., the learning rate scheme, the mini-batch size, the total epochs in the standard training procedure are all kept), but only with a negligible overhead for updating soft labels.
\par

Our main contributions are summarized as follows:
\begin{itemize}
\item Different from the current standard training procedure which discards all the learned information in the past epochs except the learned weights, we propose to learn with retrospection (LWR) to make use of the learned information in the past epochs.
LWR uses the learned instance-to-class similarities as soft labels to supervise the training in the subsequent epochs, thus alleviating overconfidence or overfitting.
\item Extensive experiments on several benchmark datasets demonstrate that LWR significantly improves accuracies, robustness, and calibration of various modern networks, and outperforms the state-of-the-art label smoothing based approaches.
\end{itemize}

\section{Related Work}

\subsection{Label Smoothing}
\cite{szegedy2016rethinking} propose the label smoothing regularizer (LSR) that utilizes the weighted average of one-hot labels and the uniform distribution as soft labels, and successfully uses it to improve the performance of the Inception architecture on image classification.
Ever since then, many advanced image classification approaches \cite{zoph2018learning,real2019regularized,huang2019gpipe} have incorporated LSR into training procedures.
Besides image classification, label smoothing has also been used in speech recognition to reduce the word error rate on the WSJ dataset \cite{chorowski2016towards}.
Moreover, in machine translation, \cite{vaswani2017attention} show that label smoothing is able to improve the BLEU score but with a reduction in perplexity.
\cite{muller2019does} empirically show that LSR is also able to improve the calibration of DNNs.
\cite{pereyra2017regularizing} propose to smooth the output distribution of a DNN by penalizing low entropy predictions, which obtains consistent performance improvements across various tasks.
Recently, \cite{yuan2020revisiting} propose a new manually designed label smoothing regularizer named TF-Reg which is developed from LSR but outperforms LSR.
These label smoothing techniques can be considered as assigning the same small probability to the non-ground-truth classes, thus mitigating the overconfidence issue of DNNs.
However, these approaches cannot fully utilize the advantages of label smoothing as they do not
take into account class similarities.

\subsection{Knowledge Distillation}
KD \cite{hinton2015distilling} is able to overcome the disadvantages of the manually designed label smoothing regularizers by taking advantage of the soft labels generated by a pretrained teacher network to train a student.
However, the training cost of KD is several times of those of using manually designed regularizers, since KD needs to first train a powerful teacher, and does inference twice (i.e., once for the teacher and once for the student) for each training sample in each iteration when training the student network.
To reduce the cost of training a powerful teacher which is usually larger than the student, many self-distillation approaches including but not limited to \cite{xu2019data,zhang2019your,yang2019snapshot,yang2019training,furlanello2018born,bagherinezhad2018label,yun2020regularizing} have been proposed.
\cite{zhang2019your} propose to generate soft labels by adding additional basic blocks or layers to the shallow layers, which improves the performance but has a large computation and memory overhead.
Born-again networks \cite{furlanello2018born} and label-refine networks \cite{bagherinezhad2018label} are based on the same idea but from different perspectives.
These two approaches train a network in many generations and use the network in the $(i-1)$th generation as the teacher to train the network in the $i$th generation.
In other words, these approaches do not pretrain a large teacher, but instead pretrain the network itself as its own teacher, and this process can be repeated many times.
%These approaches pretrain the target network many times as its own teacher instead of pretraining a large teacher network.
They still have a large training overhead as they need to train a network many times.
SD \cite{yang2019snapshot} borrows the idea from \cite{huang2017snapshot} by relying on a cyclic learning rate schedule \cite{loshchilov2016sgdr} to train DNNs in many mini-generations.
Consequently, SD cannot use the optimal training hyper-parameters (e.g., training epochs and learning rate schemes) already searched in the standard training procedure for modern network architectures such as ResNet \cite{he2016deep} and VGG \cite{simonyan2014very}.
Moreover, this cyclic learning rate scheme also causes that SD cannot update the supervision information frequently.
We also find that SD is prone to severe underconfidence.\par
Note that in this paper, we aim to propose a framework to improve the DNN performance with comparable training and inference costs to the standard training procedure.
Thus, we do not compare the proposed method with those approaches (e.g., born-again and label-refine networks) whose training or inference cost is several times of ours.

\section{Framework}
In this section, we introduce LWR which trains a DNN with the assistance of itself in the past.
To illustrate the connection between LWR and the standard training procedure, we first review the standard process for training DNNs on classification.
Then we make further derivations of the standard process to show how to train DNNs with retrospection.

\iffalse

\subsection{Formulation of Training Process on Classification}
Given a training data set $\mathcal D$ = $\{X, Y\}$ = $\{(x_i, y_i)\}_{i=1}^{i=N}$ where $x_i$ is an input sample, $y_i$ is the corresponding one-hot label, and $N$ is the number of training samples in $D$, a DNN $f$ with parameters $\Theta$ is trained on $D$.
In standard training procedure, the cross-entropy loss between DNN outputs and one-labels is minimized by a gradient descent based optimizer such as SGD with momentum or Adam.
Parameters $\Theta$ are updated iteratively based on the average gradient on a mini-batch of training samples.
Suppose that the number of the total training epochs and the mini-batch size are $K$ and $S$, respectively.
In each epoch, every data sample is used once so that the number of iterations in each epoch is $\lceil\frac{N}{S}\rceil$.
Parameters $\Theta$ are updated in each iteration based on the average gradient on a mini-batch of training samples.
With the above notations, we show LWR below.
\fi
\subsection{Standard Training Procedure}
Given a training data set $\mathcal D$ = $(X, Y)$ = $\{(x_i, y_i)\}_{i=1}^{N}$ where $x_i$ is an input sample; $y_i$ is the corresponding one-hot label; and $N$ is the total number of training samples in $\mathcal D$, a DNN $f$ with parameters $\Theta$ is trained on $\mathcal D$.
In the standard training procedure, the cross-entropy loss between DNN outputs and one-labels is minimized by a gradient descent based optimizer such as SGD with momentum or Adam \cite{kingma2014adam}.
The DNN is trained for many epochs to well fit the data, where each epoch means going through all the samples in the training set once.
Suppose that the number of total training epochs and the mini-batch size are $M$ and $B$, respectively.
The total number of iterations in each epoch is $\lceil\frac{N}{B}\rceil$.
In each iteration, a mini-batch of training samples ($\mathbf{x}$, $\mathbf{y}$) are sampled from $\mathcal D$ and then are fed into DNN $f$:
\begin{equation}
\label{1}
\mathbf{z} = f(\Theta, \mathbf{x})
\end{equation}
Note that $\mathbf{z}$ are the output logits before softmax.
Then the cross-entropy loss between the output logits and the one-hot labels are computed:
\begin{equation}
\label{2}
\mathcal{L}_{CE} = H(\sigma(\mathbf{z}), \mathbf{y} )
\end{equation}
where $H$ is cross-entropy and $\sigma(.)$ denotes softmax.
Then parameters $\Theta$ are updated once in this iteration based on the gradients of $\mathcal{L}_{CE}$ with respect to $\Theta$.

\par

From the beginning to the end of the training process, more and more information is learned with the observation that the training and validation accuracies are higher and higher.
It indicates that even in the early iterations, there is still some useful information that has been learned.
However, we notice that all the information learned in the previous iterations except the values of $\Theta$ is discarded in the standard training process.
In light of this, we propose to learn with retrospection to make use of the learned information in the previous iterations to assist the subsequent training.
An interesting problem is how to effectively use the learned information to assist the training in the subsequent iterations.

%and the one-hot labels are used in each iteration of the whole training without using the learned information to do any changes.

\subsection{Learning with Retrospection}
We propose LWR which takes advantage of the training logits in the previous epochs to generate soft labels to guide the training in the subsequent epochs.
Thus, besides the one-hot label, each training data sample also has a soft label generated during the training process.\par

\subsubsection{Making Use of Training Logits Instead of Discarding}
In each iteration of the standard training procedure, training logits $\mathbf{z}$ of a mini-batch of samples are only used to compute cross-entropy loss (\ref{2}) and then are discarded.
We argue that the training logits contain instance-to-class similarities which may be useful for subsequent training.
We take advantage of the training logits by using the softmax function with a temperature to generate soft labels:
\begin{equation}
\label{3}
\mathbf{s} = \sigma(\mathbf{z}/\tau) = \frac{exp(\mathbf z/\tau)}{\sum_j  exp(\mathbf z[j]/\tau)}
\end{equation}
where $\tau$ is a temperature to soften the logits and $z[j]$ are the logits corresponding to the $j$th class.
As every training sample is processed once in each epoch, we can obtain the soft labels for all the training samples in each epoch.
%Note that the soft labels do not necessarily need to store at GPU memory, and they can stored in the regular memory or disks.
%We denote the soft labels generated in the $i$ epoch by $\mathbf{s}_i$

\begin{figure}[t]
\setlength{\abovecaptionskip}{0.3cm}
\setlength{\belowcaptionskip}{-0.2cm}
\centering
     \includegraphics[width=0.47\textwidth]{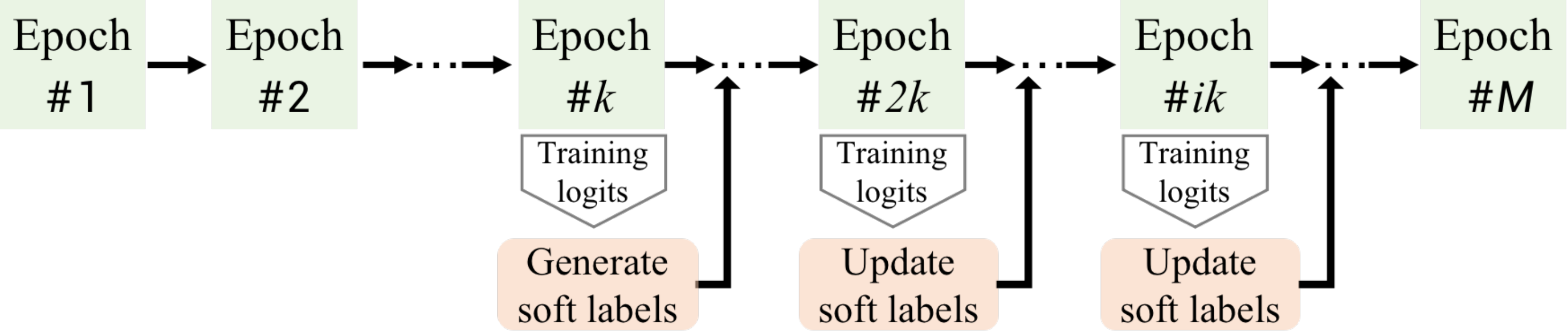}
     \caption{Framework of LWR}
     \label{f2}
\end{figure} 

\subsubsection{Training DNNs with LWR}
As shown in Figure \ref{f2}, we update the soft labels once every $k$ epochs.
In the first $k$ epochs, there are no soft labels and we just minimize the regular cross-entropy loss (\ref{2}).
After that, the training is supervised by both one-hot labels and soft labels.
We denote the soft labels generated in the $(i\times k)$th epoch by $\mathbf s_{ik}$.
$\mathbf s_{ik}$ are used to guide the training from the $(i\times k+1)$th epoch to the $((i+1)\times k)$th epoch with the following training objective:
\begin{equation}
\label{4}
\mathcal{L}_{LWR} =\alpha H(\sigma(\mathbf{z}), \mathbf{y} ) + \beta \tau^2 K(\sigma(\mathbf{z}/\tau), \mathbf{s}_{ik} )
\end{equation}
where $K(.)$ denotes KL-divergence; $\alpha$ and $\beta$ are two balancing weights.
Note that the gradients do not flow through $\mathbf{s}_{ik}$, since they are just the records of the previous epochs.
As the soft labels are more and more accurate as the training process progresses, $\beta$ should be larger and larger and $\alpha$ should be smaller and smaller from the beginning to the end of the training process.
Based on this idea, in most cases, we simply set $\alpha$ and $\beta$ to $(1-0.9 \times\frac{i\times k}{M})$ and $(0.9\times \frac{i\times k}{M})$, respectively, where $M$ is the total number of training epochs.
Thus, when LWR is introduced, almost only two hyperparameters, i.e., temperature $\tau$ and updating interval $k$, need to be tuned as LWR can use the training hyperparameters (i.e., total number of training epochs $M$, the learning rate scheme, and mini-batch size $B$) of the standard training procedure.
The training overhead of LWR is the cost of storing a soft label for each image, which is negligible as these soft labels do not need to be stored in the GPU memory.\par
The implementation-level description of LWR is summarized in Algorithm \ref{g1}.

    \begin{algorithm}[!t]  
        \caption{LWR}  
        \label{g1}
        \begin{algorithmic}[1] 
         \renewcommand{\algorithmicrequire}{\textbf{Input:}} 
          \renewcommand{\algorithmicensure}{\textbf{Output:}}
            \Require Training data $\mathcal{D}$, DNN $f$ with parameter $\Theta$
            \Ensure Optimal $\Theta$
            \For{$i = 1, 2, ..., M$ epochs} 
                 \If{$i\leq k$}
                 \State Update $\Theta$ based on (\ref{2}) by gradient descent
                  \If{$i==k$}
                 \State Generate soft labels by (\ref{3})
                  \EndIf
                 \Else
                 \State Update $\Theta$ based on (\ref{4}) by gradient descent
                 \If{$i\%k==0$}
                 \State Update soft labels by (\ref{3})
                  \EndIf   
                  \EndIf  
                \EndFor  
        \end{algorithmic}  
    \end{algorithm}

\subsection{Why LWR Works}%understand and improve knowledge distillation
In this part, we provide an analysis of why LWR works.
\subsubsection{Benefiting from Label Smoothing}
LSR smooths one-hot label $y$ by using a weighted average of the one-hot label and a uniform distribution, i.e., $y_{LSR}=(1-\epsilon)*y+\epsilon/C$ where $\epsilon$ is a small factor and usually set to 0.1, and $C$ is the total number of classes.
As shown in the existing literature \cite{muller2019does}, LSR mitigates over-confidence of DNNs and improves DNN generalization and calibration.
We notice that LSR is a special case of LWR.
%The training loss using $y_{LSR}$ can be decomposed to two parts
The training loss of LSR can be decomposed into two parts:
\begin{equation}
\label{5}
H(\sigma(\mathbf{z}), y_{LSR} ) = (1-\epsilon)H(\sigma(\mathbf{z}), y)+\epsilon H(\sigma(\mathbf{z}), \frac{1}{C})
\end{equation}
It is observed that (\ref{5}) shares a similar form to (\ref{4}), and $K(\sigma(\mathbf{z}/\tau), \mathbf{s}_{ik})$ in (\ref{4}) is equivalent to $H(\sigma(\mathbf{z}/\tau)), \mathbf{s}_{ik})$ plus a constant (i.e., the entropy of soft labels $\mathbf{s}_{ik}$).
When $\tau$ is set to 1 and the learned soft label $\mathbf{s}_{ik}$ follows a uniform distribution, LWR is equivalent to LSR.
%In other words, 
Thus, LWR is an adaptive version of LSR, suggesting that it should inherit the advantages of LSR, such as better generalization and calibration.

\begin{table*}[!t]
\setlength{\abovecaptionskip}{0.1cm}
\setlength{\belowcaptionskip}{0.1cm}
\centering
\caption{Test Accuracies (\%) on CIFAR-100 and Tiny ImageNet. $\uparrow$ denotes the absolute improvement over the standard training procedure (i.e., STD).}
\label{m1}
\resizebox{\textwidth}{!}{%
\begin{tabular}{lllllll}
\toprule
                     & \multicolumn{1}{l}{} & ResNet-56     & WRN-16-4      & ShuffleV2     & VGG-16   & PreAct ResNet-18     \\\midrule
\multirow{7}{*}{\begin{tabular}[c]{@{}l@{}}CIFAR \\ -100\end{tabular}} &STD           & 72.00$\pm$0.16 & 76.43$\pm$0.16 & 71.12$\pm$0.39 & 74.16$\pm$0.28  & 77.31$\pm$0.33 \\
                     &LSR           & 72.05$\pm$0.16 & 76.45$\pm$0.10 & 71.90$\pm$0.13 & 74.75$\pm$0.12 & 78.24$\pm$0.16 \\
                     &Max-Entropy  & 72.03$\pm$0.19 & 76.47$\pm$0.15 & 71.23$\pm$0.22 & 74.11$\pm$0.13  & 77.65$\pm$0.35   \\
                     &SD            & 72.22$\pm$0.20 & 77.00$\pm$0.38 & 68.70$\pm$0.51 & 74.28$\pm$0.35  & 78.31$\pm$0.24  \\
                     &CS-KD       & 72.05$\pm$0.19 & 76.28$\pm$0.11 &71.43$\pm$0.28  & 74.61$\pm$0.12  & 78.01$\pm$0.13   \\
                     &TF-Reg       & 72.11$\pm$0.29 & 76.52$\pm$0.28 & 72.09$\pm$0.34 & 74.69$\pm$0.18  & 77.36$\pm$0.23   \\
                     &LWR (Ours)   & \textbf{74.25$\pm$0.29 ($\uparrow$ 2.25)} & \textbf{77.88$\pm$0.27 ($\uparrow$ 1.45)} & \textbf{73.53$\pm$0.33 ($\uparrow$ 2.41)} & \textbf{75.24$\pm$0.09 ($\uparrow$ 1.08)} & \textbf{79.73$\pm$0.32 ($\uparrow$ 2.42)}   \\ \midrule \midrule
            %& \multicolumn{1}{l}{} & ResNet-56          & WRN-16-4         & ShuffleV2     & VGG-16        & PreAct ResNet-18    \\
\multirow{7}{*}{\begin{tabular}[c]{@{}l@{}}Tiny \\ImageNet\end{tabular}}  &STD         & 56.31$\pm$0.05    & 59.48$\pm$0.16 & 60.80$\pm$0.25 & 62.39$\pm$0.45 & 65.57$\pm$0.16 \\
                    &LSR          & 56.57$\pm$0.38    & 59.33$\pm$0.16 & 61.85$\pm$0.27 & 63.82$\pm$0.03 & 64.91$\pm$0.08 \\
                   &Max-Entropy & 56.80$\pm$0.41    & 59.06$\pm$0.19 & 61.56$\pm$0.40 & 62.99$\pm$0.12 & 65.25$\pm$0.16 \\
                &SD          & 57.52$\pm$0.17.    &59.68$\pm$0.13 &60.79$\pm$0.15 &63.14$\pm$0.16 &65.87$\pm$0.28  \\
                &CS-KD      & 56.21$\pm$0.41    &59.92$\pm$0.29   &61.66$\pm$0.34 & 62.87$\pm$0.20 & 64.29$\pm$0.25 \\
                &TF-Reg      & 56.43$\pm$0.19    & 59.41$\pm$0.10 & 61.55$\pm$0.42 & 62.95$\pm$0.06 & 64.88$\pm$0.47 \\
                &LWR (Ours)  & \textbf{57.95$\pm$0.25 ($\uparrow$ 1.64)}    & \textbf{61.22$\pm$0.29 ($\uparrow$ 1.74)} & \textbf{62.06$\pm$0.29 ($\uparrow$1.26)} & \textbf{64.42$\pm$0.06 ($\uparrow$ 2.03 )} & \textbf{66.40$\pm$0.12 ($\uparrow$ 0.83)} \\ \bottomrule
\end{tabular}
}
\vspace{-0cm}
\end{table*}

% Please add the following required packages to your document preamble:
% \usepackage{multirow}
\begin{table*}[!t]
\setlength{\abovecaptionskip}{0.1cm}
\setlength{\belowcaptionskip}{0.1cm}
\centering
\caption{Test Accuracies (\%) on fine-grained classification datasets. $\uparrow$ denotes the absolute improvement over the STD procedure.}
\label{m3}
\resizebox{\textwidth}{!}{%
\begin{tabular}{llllll}
\toprule
                               & \multicolumn{1}{l}{} & ResNet-10      & ResNet-18      & MobileNetV2    & ShuffleNetV2   \\ \midrule
\multirow{7}{*}{CUB}           & STD                  & 58.96$\pm$0.12 & 61.09$\pm$0.49 & 67.20$\pm$0.21 & 61.67$\pm$0.85 \\
                               & LSR                  & 59.31$\pm$0.21 & 63.57$\pm$0.50 & 67.97$\pm$0.43 & 62.66$\pm$0.11 \\
                               & Max-Entropy          & 59.00$\pm$0.30 & 61.23$\pm$0.37 &66.56$\pm$0.30  &61.10$\pm$0.15  \\
                               & SD                   & 59.28$\pm$0.34 & 64.19$\pm$0.24 & 68.15$\pm$0.32 & 63.99$\pm$0.29 \\
                               & CS-KD               &60.70$\pm$0.21  &64.57$\pm$0.29  &67.48$\pm$0.32  &63.32$\pm$0.25   \\
                               & TF-Reg               & 58.84$\pm$0.60 &62.04$\pm$0.28 &67.20$\pm$0.43&61.19$\pm$0.58                \\
                               & LWR (Ours)           & \textbf{63.14$\pm$0.17 ($\uparrow$ 4.18)} & \textbf{66.47$\pm$0.37 ($\uparrow$ 5.38)} & \textbf{69.00$\pm$0.41 ($\uparrow$ 1.80)} & \textbf{64.37$\pm$0.52 ($\uparrow$ 2.70)} \\ \midrule \midrule 
\multirow{7}{*}{Stanford Dogs} & STD                  &63.91$\pm$0.25  &66.56$\pm$0.28  &68.05$\pm$0.26 &66.08$\pm$0.32  \\
                               & LSR                  &63.36$\pm$0.03  &67.12$\pm$0.86  &69.13$\pm$0.09 &66.90$\pm$0.34 \\
                               & Max-Entropy          &63.94$\pm$0.30  &66.42$\pm$0.50  &67.97$\pm$0.30  &66.25$\pm$0.60    \\
                               & SD                   &64.65$\pm$0.36  &68.79$\pm$0.06  &70.26$\pm$0.35  &67.30$\pm$0.26       \\
                               
                               & CS-KD               &64.91$\pm$0.26  &69.17$\pm$0.19  &68.73$\pm$0.25  &66.75$\pm$0.31   \\
                               
                               & TF-Reg               &63.72$\pm$0.44  &66.53$\pm$0.50  &68.36$\pm$0.26  &66.63$\pm$0.26   \\
                               & LWR (Ours)           & \textbf{66.28$\pm$0.15 ($\uparrow$ 2.37)} &\textbf{69.84$\pm$0.45 ($\uparrow$ 3.28)} &\textbf{70.45$\pm$0.12 ($\uparrow$ 2.40)} &\textbf{67.39$\pm$0.42 ($\uparrow$ 1.31)} \\ \midrule \midrule  
\multirow{7}{*}{FGVC-Aircraft} & STD                  &73.89$\pm$0.25  &79.58$\pm$0.25  &83.01$\pm$0.30   & 78.00$\pm$0.45     \\
                               & LSR                  &74.52$\pm$0.18  &80.91$\pm$0.28    &83.88$\pm$0.10   &78.40$\pm$0.94    \\
                               & Max-Entropy          &73.39$\pm$0.09  &79.59$\pm$0.41   &82.81$\pm$0.45    &78.20$\pm$0.25   \\
                               & SD                   &74.98$\pm$0.38  &80.55$\pm$0.80   &83.36$\pm$0.13  &78.09$\pm$0.34    \\
                               & CS-KD               &74.95$\pm$0.40  &79.72$\pm$0.19 &80.62$\pm$0.38  &77.89$\pm$1.55   \\
                               & TF-Reg               &73.69$\pm$0.38  &80.12$\pm$0.33 &83.39$\pm$0.11  &78.50$\pm$0.31   \\
                               & LWR (Ours)           &\textbf{76.41$\pm$0.26 ($\uparrow$ 2.52)}   &\textbf{81.25$\pm$0.05 ($\uparrow$ 1.67)}       &\textbf{84.56$\pm$0.54 ($\uparrow$ 1.55)}   & \textbf{78.59$\pm$0.37 ($\uparrow$ 0.59)}               \\ \bottomrule
\end{tabular}
}
\end{table*}

\iffalse
\begin{table}[!t]
\setlength{\abovecaptionskip}{0.1cm}
\setlength{\belowcaptionskip}{0.1cm}
\centering
\caption{TOP-1 Accuracies (\%) on ImageNet.}
\label{mm}
\begin{tabular}{lllll}
\toprule
           &STD   &LSR    &CS-KD &LWR           \\\midrule
ResNet-18  & 69.6  &70.0  &70.0 &\textbf{70.3 $\uparrow$(0.7)}\\
ResNet-50  & 75.9  &76.3  &76.4 &\textbf{76.5 $\uparrow$(0.6)}   \\ \bottomrule                               
\end{tabular}
\vspace{-0.5cm}
\end{table}
\fi

\begin{table}[!t]
\setlength{\abovecaptionskip}{0.1cm}
\setlength{\belowcaptionskip}{0.1cm}
\centering
\caption{Test Accuracies (\%) on Tabular Datasets}
\label{mtabular}
\resizebox{0.45\textwidth}{!}{%
\begin{tabular}{llll}
\toprule
      &Abalone           &Arcene       &Iris        \\\midrule
STD   &25.75$\pm$0.26   &83.00$\pm$2.16 &90.00$\pm$2.72\\
LSR   &26.75$\pm$0.54   &81.00$\pm$4.55 &92.22$\pm$1.57    \\ 
Max-H &27.15$\pm$1.17   &79.67$\pm$3.77 &94.44$\pm$1.57 \\
SD  &25.79$\pm$0.91     &81.00$\pm$1.63    &93.33$\pm$2.72 \\
CS-KD &27.70$\pm$1.08    &83.67$\pm$0.94    & 94.44$\pm$1.57\\ 
TF-Reg &25.87$\pm$0.52 &80.00$\pm$2.16    &90.00$\pm$2.72 \\
LWR (Ours) &\textbf{31.86$\pm$0.51} &\textbf{85.33$\pm$1.25}   & \textbf{95.56$\pm$1.57}  \\ \bottomrule  
\end{tabular}
}
\vspace{-0.5cm}
\end{table}

%$\uparrow$(6.11). $\uparrow$(2.33)  $\uparrow$(5.56)

\subsubsection{Benefiting from Instance-level Class Similarities}
One-hot labels set the probability for the ground-truth class to 1 while the probabilities for the other classes are all set to 0.
This may cause overfitting or overconfidence issues especially when the training samples in different classes share visual or semantic similarities.
Maximizing the ground-truth probability while preserving small probabilities for the other classes may mitigate this issue. 
It is principled to use class similarities as the soft labels which assign corresponding probabilities to different classes.
Note that class similarities may not well represent instance-to-class similarities because different training samples in the same class may be close to different classification boundaries.
For example, some cats are more visually similar to dogs while some other cats are more similar to tigers.
This kind of instance-level class similarities is learned gradually during the training process.
Motivated by this observation, LWR takes advantage of the training logits to generate instance-level soft labels.
Therefore, LWR also benefits from the learned instance-to-class similarities during the training process.%, thus improving the performance.

%\subsubsection{Benefiting from Consistency}
%Modern DNNs are typically trained with data augmentation.
%Thus, the samples used in each epoch are randomly augmented samples (e.g., randomly cropping and flipping) from the original data.
%This means that the samples used in different epochs are not identical but are generated from the same samples.
%As LWR uses the soft labels generated from previous epochs as targets to guiding future training, LWR implicitly forces the consistent outputs across different augmented versions of the data.

\section{Experiments}
%We evaluate LWR on several benchmark datasets with various modern network architectures.

\subsection{Experimental Setup}

\subsubsection{Datasets}
%ImageNet \cite{deng2009imagenet}
We report the results on several benchmark datasets, i.e., CIFAR-10 \cite{krizhevsky2009learning}, CIFAR-100 \cite{krizhevsky2009learning}, Tiny ImageNet \footnote{https://tiny-imagenet.herokuapp.com}, CUB-200-2011 \cite{wah2011caltech}, Stanford Dogs \cite{khosla2011novel}, FGVC-Aircraft \cite{maji13fine-grained}, Abalone \cite{Dua:2019}, Arcene \cite{Dua:2019}, and Iris \cite{Dua:2019}.
CIFAR-10 is a 10-class image classification dataset, containing 50,000 training images and 10,000 test images.
CIFAR-100 has similar images to those in CIFAR-10, but has 100 classes. 
Tiny ImageNet, i.e., a subset of ImageNet, has 200 classes, containing 100,000 training images and 10,000 test images.
%ImageNet is a large-scale image classification dataset with 1000 classes, containing 1.28 million training images and 50,000 validation images.
CUB-200-2011, Stanford Dogs, and FGVC-Aircraft are three fine-grained classification datasets, containing 11,788 images of 200 bird species, 22,000 images of 120 breeds of dogs, and 10,200 images of 102 different aircraft model variants, respectively.
%MIT67 contains 15,620 images of 67 indoor categories. contains 
Abalone, Arcene, and Iris are three tabular datasets which are randomly drawn from UCI datasets \cite{Dua:2019}.
We follow the default training and test splits.
%The images in each dataset are preprocessed by subtracting the mean of the dataset and being divided by the standard deviation.
We use the standard data augmentation strategy for image datasets, i.e., randomly flipping horizontally, padding, and then randomly cropping.
More details are presented in the Appendix.

\par

\subsubsection{Architectures}
To check whether LWR is able to work on different network architectures, we adopt a variety of modern architectures including ResNet \cite{he2016deep}, PreAct ResNet \cite{he2016identity}, VGG \cite{simonyan2014very}, WRN \cite{Zagoruyko2016WRN}, MobileNet \cite{sandler2018mobilenetv2}, and ShuffleNet \cite{ma2018shufflenet}.
\par

\subsubsection{Competitors}
We compare LWR with the standard training procedure and label smoothing based methods including cost-comparable self-distillation methods: (1) STD: STD trains a DNN by minimizing the regular cross-entropy loss between output logits and one-hot labels; (2) LSR \cite{muller2019does,szegedy2016rethinking}: LSR uses the weighted average of one-hot labels and a uniform distribution as targets to train a DNN.
(3) Max-Entropy \cite{pereyra2017regularizing}: Max-Entropy smooths the DNN output by maximizing its entropy.
(4) SD \cite{yang2019snapshot}: SD is a self-distillation method that relies on a periodic learning rate scheme to train a DNN.
(5) CS-KD \cite{yun2020regularizing}: CS-KD distills the predictive distribution of different samples from the same class.
(6) TF-Reg \cite{yuan2020revisiting}: TF-Reg modifies LSR to generate more accurate soft labels.

\subsubsection{Hyperparameters}
Following the standard training procedure for modern DNNs, we have trained all the networks for 200 epochs with optimizer SGD with momentum 0.9 and weight decay 5e-4 on CIFAR, CUB-200-2011, Stanford Dogs, and FGVC-Aircraft, 120 epochs for Tiny ImageNet.
%, and 90 for ImageNet with weight decay 1e-4.
%The mini-batch size is set to 128 for CIFAR and Tiny ImageNet, and 32 for the other datasets.
%The initial learning rate is set to 0.1 for all the networks except for ShuffleNet and MobileNet with 0.05 on CIFAR and Tiny ImageNet, and 0.025 on CUB-200-2011, Stanford Dogs, and FGVC-Aircraft.
%For ResNet, PreAct ResNet, MobileNet, and ShuffleNet, we follow the widely used scheme and divide the learning by 10 after 50\% and 75\% of the total training epochs.
%For WRN, we follow the original paper and divide the learning rate by 5 after 60, 120, and 180 epochs on all the datasets except Tiny ImageNet on which the learning rate is divided by 5 after 30, 60, and 90 epochs.
%For VGG, we follow the commonly used strategy and divide the learning rate by 2 every 20 epochs.
More implementation details are reported in Appendix.
%$\tau$ is set to 5 for all the networks except 50 and 10 for MobileNet and ShuffleNet on FGVC-Aircraft, respectively.
%For ResNet, PreAct ResNet, MobileNet, and ShuffleNet, $k$ is set to 5 on all the datasets except 3 on Tiny ImageNet and 3 for MobileNet on FGVC-Aircraft.
%For VGG, $k$ is set 50 on all the datasets except 30 on Tiny ImageNet.
%For WRN, $k$ is set to 1.
%Based on sampling the training data, if noise rate on CIFAR-10 is 0.2, \alpha and \beta are set to 0.1+(1-ki/100)*0.9 and  (ki/100*0.9, respectively and when the learning rate decreases,  \alpha and \beta are set to 0.1 and 0.9, respectively.
%If noise rate on CIFAR-10 is greater than 0.2, \alpha and \beta are set to 1-ki/100 and  ki/100, respectively and when the learning rate decreases,  \alpha and \beta are set to 0.0 and 1.0, respectively.
For all the competitors, we report the author-reported results or use author-provided codes and the optimal hyper-parameters from the original papers if they are publicly available.
Otherwise, we use our implementation.
We report the test accuracy in the last epoch unless otherwise specified.
All the results below are reported based on 3 runs.

\begin{table*}[!t]
\centering
\setlength{\abovecaptionskip}{0.1cm}
\setlength{\belowcaptionskip}{0.1cm}
\caption{ECE (\%) Results. For ECE, the lower is better. $\downarrow$ denotes the absolute ECE reduction below the STD training procedure.}
\label{m4}
\begin{tabular}{lll|ll}
\toprule
            & \multicolumn{2}{c|}{CIFAR-100}    & \multicolumn{2}{c}{Tiny ImageNet} \\
            & ShuffleNetV2   & Preact ResNet-18 & ShuffleNetV2   & Preact ResNet-18 \\ \midrule
STD         & 12.60$\pm$0.16 & 7.52$\pm$0.30    & 10.12$\pm$0.22 & 11.04$\pm$0.11   \\
LSR         & 3.16$\pm$0.47  & 10.81$\pm$0.44   &  \textbf{2.89$\pm$0.15}  & 8.33$\pm$0.43    \\
Max-Entropy & 13.87$\pm$0.24 & 8.41$\pm$0.14    & 11.88$\pm$0.47 & 13.43$\pm$0.11   \\
SD          & 28.63$\pm$0.38 & 32.74$\pm$0.18   & 25.34$\pm$0.50 & 31.03$\pm$0.21   \\
CS-KD      & 8.84$\pm$0.19   & 4.69$\pm$0.56   & 5.33$\pm$0.31  & 6.66$\pm$0.43   \\
TF-Reg      & 3.25$\pm$0.20   & 10.76$\pm$0.14   & 8.16$\pm$0.43  & 9.13$\pm$0.42    \\
LWR (Ours)  & \textbf{2.87$\pm$0.06 ($\downarrow$ 9.73)}  &\textbf{3.53$\pm$0.13 ($\downarrow$ 3.99)}    & 4.30$\pm$0.33 ($\downarrow$ 5.82)  &  \textbf{1.58$\pm$0.32 ($\downarrow$ 9.46)}   \\ \bottomrule
\end{tabular}
\end{table*}

\begin{figure*}[]
  %\vskip 0.2in
     \begin{minipage}{0.23\textwidth}
     \centering
     \includegraphics[height=3.9cm]{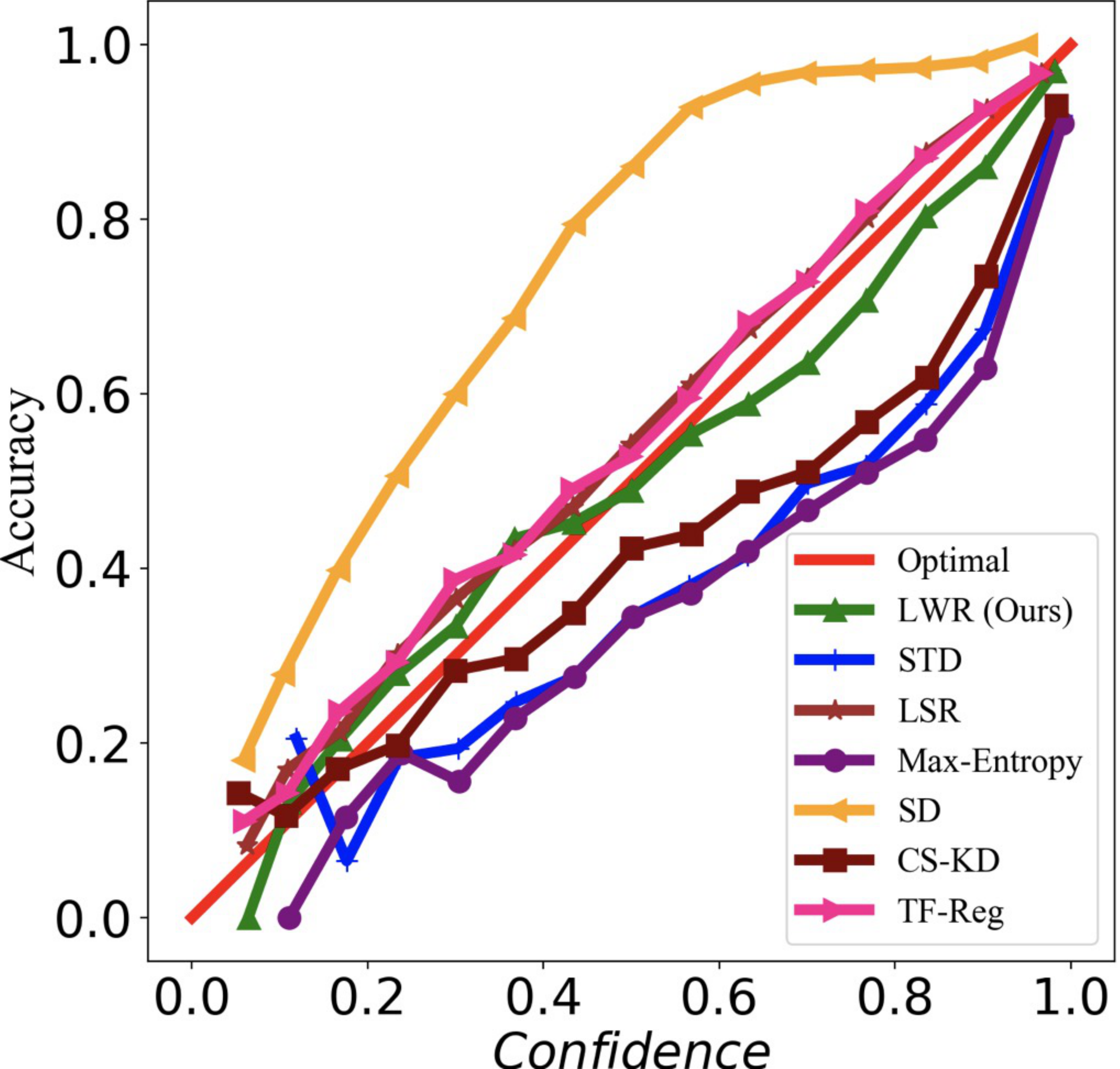}
     \caption{ShuffleNetV2 on CIFAR-100}
     \label{f3}
   \end{minipage}\hfill
   \begin{minipage}{0.23\textwidth}
     \centering
     \includegraphics[height=3.9cm]{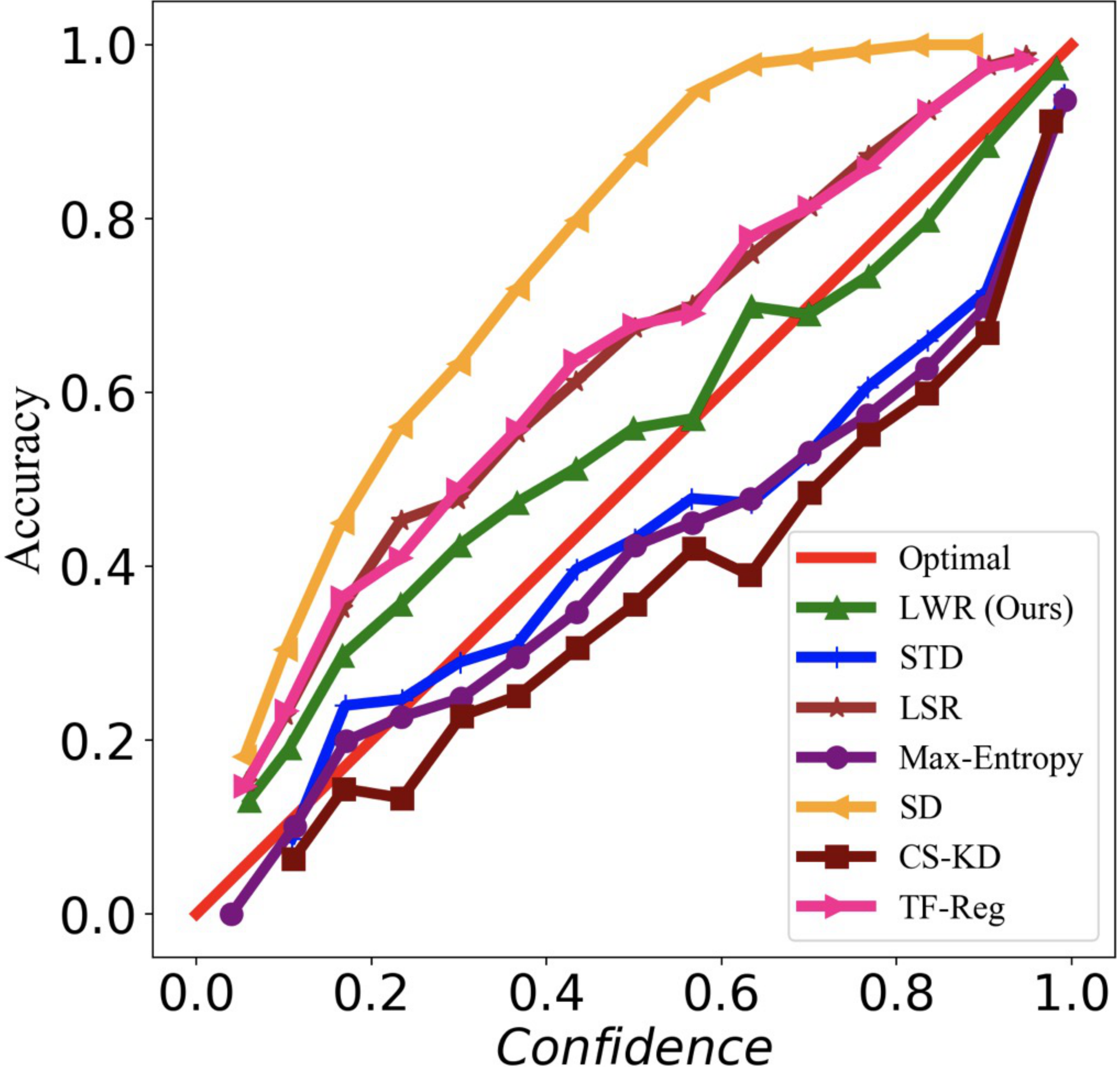}
     \caption{Preact ResNet-18 on CIFAR-100}
     \label{f4}
   \end{minipage}\hfill
   \begin{minipage}{0.23\textwidth}
     \centering
     \includegraphics[height=3.9cm]{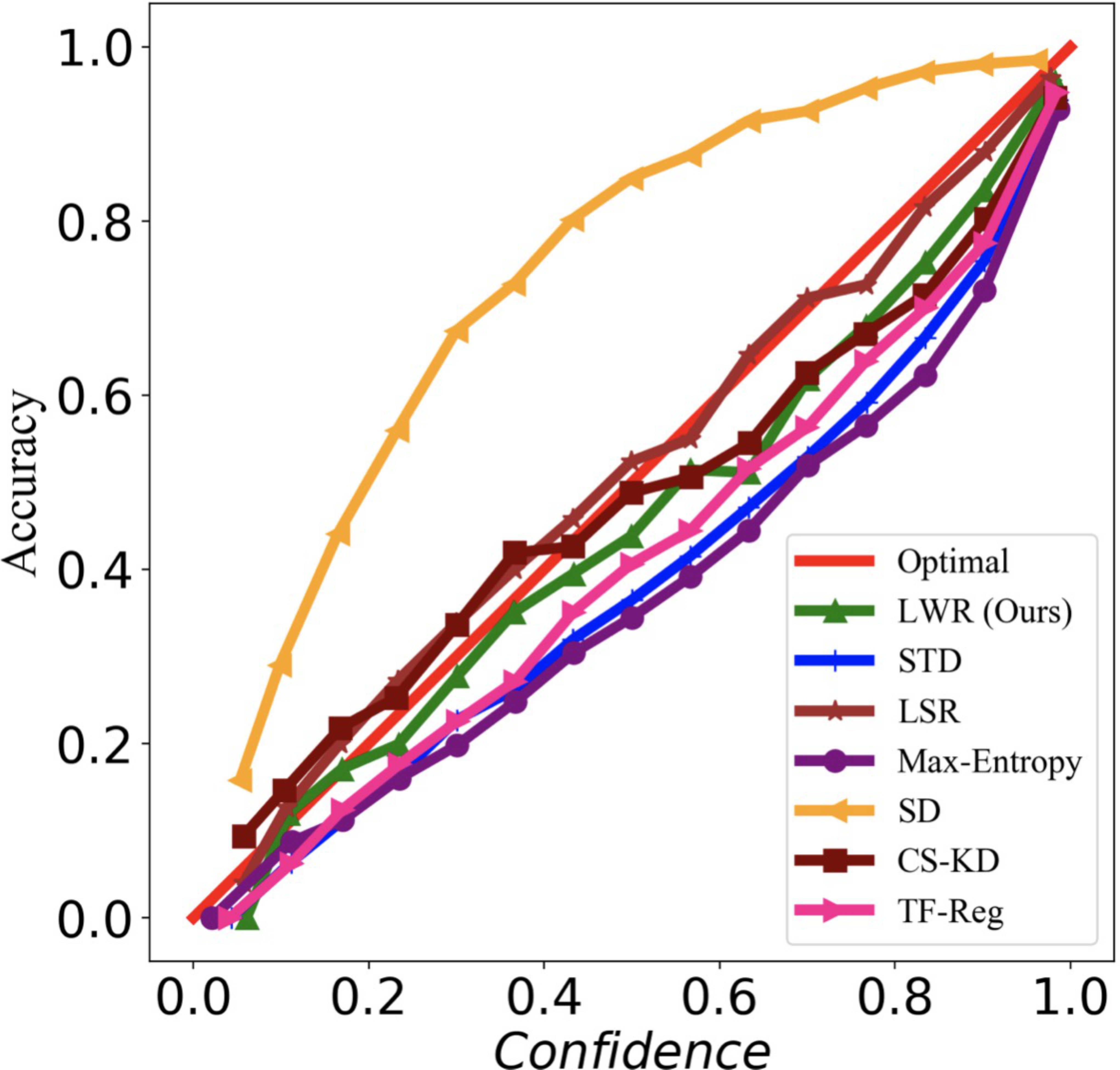}
     \caption{ShuffleNetV2 on Tiny ImageNet}
     \label{f5}
   \end{minipage}\hfill   
   \begin{minipage}{0.23\textwidth}
     \centering
     \includegraphics[height=3.9cm]{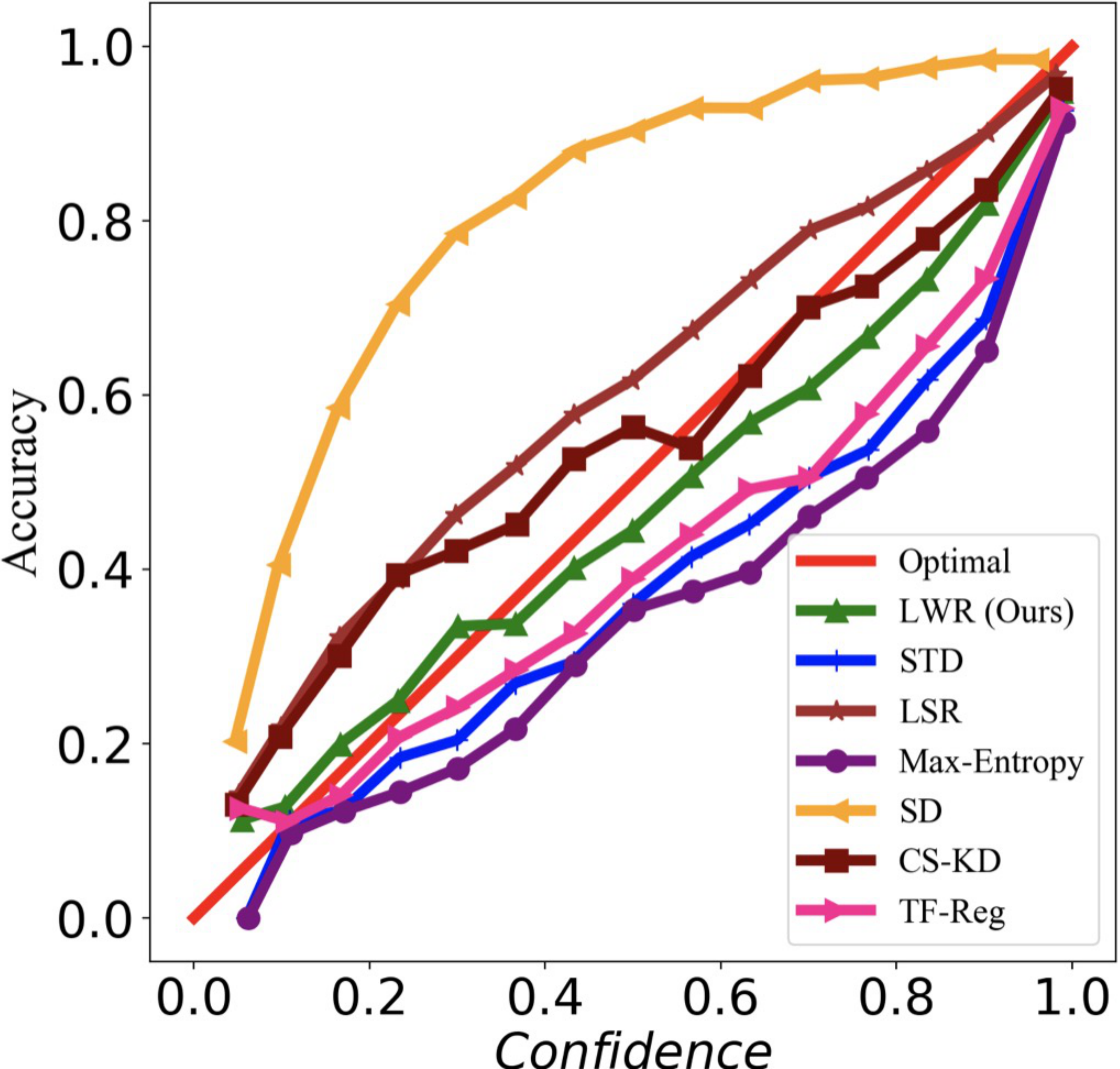}
     \caption{Preact ResNet-18 on Tiny ImageNet}
     \label{f6}
   \end{minipage}  
   \vskip -0.2in
\end{figure*}

\subsection{Classification Accuracy}
%We use CIFAR-100 and Tiny ImageNet datasets for regular classification, CUB-200-2011, Stanford Dogs, and FGVC-Aircraft datasets for fine-grained classification, Analone, Arcene, and Iris for tabular data classification.

%and ImageNet for large-scale classification.\par
\subsubsection{Regular Classification}
We use CIFAR-100 and Tiny ImageNet datasets for regular classification.
The results are reported in Table \ref{m1}.
It is observed that by simply  learning from the past, LWR improves the performances by a large margin over the standard training procedure (i.e., STD) on both datasets across various modern DNN architectures, and also outperforms all the other label smoothing based approaches significantly, which demonstrates the effectiveness of LWR.\par

\subsubsection{Fine-grained Classification}
We adpot CUB-200-2011, Stanford Dogs, and FGVC-Aircraft datasets for fine-grained classification. Table \ref{m3} reports the comparison results on these three fine-grained datasets.
LWR obtains much better performances than those of the standard training procedure and the other label smoothing based approaches.
Overall, the superiority of LWR becomes more obvious on fine-grained datasets.
This is not surprising due to the following facts: (1) fine-grained image classification contains more visually similar classes; (2) STD uses one-hot labels which are orthogonal for different classes, ignoring class similarities; (3) in contrast, LWR takes advantage of soft labels which contain instance-to-class similarities.
This also implies that the instance-to-class similarity is significantly important for fine-grained classification.\par

%To evaluate the applicability of LWR on large scale datasets, we conduce experiments on ImageNet.
%As reported in Table \ref{mm}, LWR improves more than 0.6\% of the top-1 accuracy on ImageNet across different networks without introducing any additional network parameters, which demenstrates the effectiveness of LWR on large scale datasets.
\subsubsection{Tabular Data Classification}
To evaluate LWR on non-image data, we conduct a series of experiments
on three tabular datasets which are randomly drawn from the UCI dataset.
We follow \cite{zhang2017mixup} and adopt the neural network with two hidden, fully-connected layers of 128 units.
We train it for 50 epochs with mini-batche size 16 by using Adam \cite{kingma2014adam} with default
hyper-parameters.
As shown in Table \ref{mtabular}, LWR improves 6.11\%, 2.33\%, and 5.56\% of the accuracies over the strand training procedure on the three datasets, respectively, and outperforms the other approaches significantly, which demonstrates the applicability of LWR on non-image data.

\begin{table*}[!t]
\setlength{\abovecaptionskip}{0cm}
\setlength{\belowcaptionskip}{0.0cm}
\caption{Robustness Results with Preact ResNet-18 on CIFAR-10 with different levels of noise}
\centering
\label{m5}
%\resizebox{\textwidth}{!}{%
\begin{tabular}{llllllll}
\toprule
Noise                 & Accuracy & STD                & LSR            & SD            & CS-KD   & TF-Reg        & LWR(Ours)      \\ \midrule
\multirow{2}{*}{20\%} & Last     & 83.81$\pm$0.13  & 84.21$\pm$0.41 & 90.13$\pm$0.17 &83.62$\pm$0.15  & 83.89$\pm$0.69 & \textbf{93.80$\pm$0.13 ($\uparrow$ 9.99)} \\
                      & Best     & 91.78$\pm$0.07  &91.77$\pm$0.18  & 90.72$\pm$0.21 &85.69$\pm$0.23 & 91.85$\pm$0.15 & \textbf{94.17$\pm$0.04 ($\uparrow$ 2.39)}\\ \midrule
\multirow{2}{*}{40\%} & Last     & 66.32$\pm$0.56  &66.93$\pm$0.92 & 83.22$\pm$0.74 &69.51$\pm$0.51 & 65.77$\pm$2.39 & \textbf{89.14$\pm$0.30 ($\uparrow$ 22.82)} \\
                      & Best     & 88.89$\pm$0.06 &89.41$\pm$0.25 & 86.57$\pm$0.20  &84.00$\pm$0.68 & 88.96$\pm$0.12 & \textbf{89.43$\pm$0.32 ($\uparrow$       0.54)} \\ \midrule
\multirow{2}{*}{60\%} & Last     & 45.05$\pm$0.71  &46.35$\pm$1.24  & 75.01$\pm$0.27 &49.90$\pm$0.45 & 45.01$\pm$1.05 & \textbf{86.19$\pm$0.05 ($\uparrow$ 41.14)} \\
                      & Best     & 84.38$\pm$0.14 &84.74$\pm$0.06  & 81.17$\pm$0.17 &77.39$\pm$0.98 & 84.51$\pm$0.06 & \textbf{87.27$\pm$0.08 ($\uparrow$       2.89)} \\ \midrule
\multirow{2}{*}{80\%} & Last     & 27.08$\pm$1.51  &26.18$\pm$0.49  & 65.79$\pm$0.68 & 27.82$\pm$0.51 & 26.52$\pm$0.26 & \textbf{72.27$\pm$0.19 ($\uparrow$ 45.19)} \\
                      & Best     & 73.32$\pm$0.23 &73.04$\pm$0.23  & 71.14$\pm$0.33 & 60.07$\pm$1.42 & 72.12$\pm$0.38 & \textbf{78.72$\pm$0.10 ($\uparrow$       5.40)} \\ \bottomrule
\end{tabular}%
%}
\end{table*}

\begin{table*}[t]
\setlength{\abovecaptionskip}{0cm}
\setlength{\belowcaptionskip}{0.0cm}
\centering
\caption{Effects of updating interval $k$ and temperature $\tau$ in terms of test accuracy (\%) on CIFAR-100}
\label{m6}
\begin{tabular}{llcccccc}
\toprule
                         &                  & k=1            & k=5            & k=10           & k=20           & k=50           & k=100          \\ \midrule
\multirow{5}{*}{$\tau$=5}  & PreAct ResNet-18 & 79.50$\pm$0.10 & 79.73$\pm$0.22 & 79.27$\pm$0.10 & 78.67$\pm$0.52 & 78.46$\pm$0.51 & 76.32$\pm$0.11 \\
                         & ResNet-56        & 73.40$\pm$0.31 & 74.25$\pm$0.29 & 73.87$\pm$0.15 & 73.37$\pm$0.17 & 73.53$\pm$0.27 & 72.98$\pm$0.23 \\
                         & WRN-16-4         & 77.88$\pm$0.27 & 77.64$\pm$0.39 & 77.63$\pm$0.18 & 77.60$\pm$0.22 & 77.37$\pm$0.29 & 76.60$\pm$0.09 \\
                         & ShuffleNetV2     & 73.45$\pm$0.05 & 73.53$\pm$0.33 & 73.55$\pm$0.30 & 73.33$\pm$0.11 & 73.20$\pm$0.19 & 72.50$\pm$0.03 \\
                         & VGG-16           & 74.97$\pm$0.08 & 75.05$\pm$0.07 & 74.94$\pm$0.16 & 75.24$\pm$0.11 & 75.24$\pm$0.09 & 74.13$\pm$0.27 \\ \midrule \midrule
\multirow{5}{*}{$\tau$=10} & PreAct ResNet-18 & 79.39$\pm$0.30 & 79.12$\pm$0.03 & 79.60$\pm$0.17 & 78.95$\pm$0.11 & 78.24$\pm$0.27 & 76.50$\pm$0.38 \\
                         & ResNet-56        & 73.66$\pm$0.15 & 73.40$\pm$0.30 & 73.62$\pm$0.29 & 73.65$\pm$0.17 & 73.81$\pm$0.19 & 72.62$\pm$0.40 \\
                         & WRN-16-4         & 77.17$\pm$0.14 & 77.43$\pm$0.20 & 77.23$\pm$0.12 & 77.84$\pm$0.13 & 77.38$\pm$0.33 & 76.64$\pm$0.12 \\
                         & ShuffleNetV2     & 73.29$\pm$0.20 & 72.89$\pm$0.37 & 73.55$\pm$0.24 & 73.16$\pm$0.10 & 73.57$\pm$0.34 & 72.89$\pm$0.09 \\
                         & VGG-16           & 74.78$\pm$0.36 & 74.87$\pm$0.32 & 74.88$\pm$0.14 & 75.14$\pm$0.26 & 75.27$\pm$0.28 & 74.15$\pm$0.16 \\ \bottomrule
\end{tabular}
\vspace{-0.4cm}
\end{table*}

\subsection{Calibration}
With the deployment of DNNs in high-risk domains, predictive uncertainty of DNNs is of increasing importance.
The predictive confidence of a well-calibrated classifier should be indicative of the accuracy.
%the actual likelihood of correctness.
Following the existing work \cite{guo2017calibration,muller2019does}, we use Expected Calibration Error (ECE) and Reliability Diagrams to measure the calibration effects.
Specifically, ECE is the expected difference between confidence scores (i.e., the winning softmax scores) and accuracies. 
It is calculated by partitioning
classifier predictions into $M$ bins of equal size and taking a weighted average of differences between confidence scores and accuracies in the bins:
\begin{equation}
\label{}
ECE=\sum_{m=1}^{M}\frac{\left|B_m\right|}{n}\left|acc\left(B_m\right)-conf\left(B_m\right)\right|
\end{equation}
where $n$ is the total number of samples; $B_m$ is the set of samples whose confidence scores fall into bin $m$; $\left|B_m\right|$ denotes the number of samples in $B_m$; $acc(B_m)$ is the accuracy of $B_m$; and $conf(B_m)$ is the average confidence of $B_m$.\par
Table \ref{m4} reports the ECE of different approaches.
LWR improves the calibration over the standard training procedure consistently and significantly.
As expected, in most cases, LWR and LSR perform better than the other approaches, and overall LWR outperforms all the other competitors.\par

To better understand the overconfidence or underconfidence issue of different approaches, we plot the reliability diagrams of these approaches.
The reliability diagram plots the expected accuracy as a function of the confidence.
Thus, the identity function implies the optimal calibration.
Figure \ref{f3}, Figure \ref{f4}, Figure \ref{f5}, and Figure \ref{f6} show the reliability diagrams of different approaches on CIFAT-100 and Tiny ImageNet with ShuffleNetV2 and Preact ResNet-18.
We plot the red line (i.e., the identity function) to represent the optimal calibration.
We have the following observations: (1) LWR, LSR, and TF-Reg are closer to the optimal calibration than the other approaches, which demonstrates the effectiveness of LWR on calibration; (2) STD and Max-Entropy are prone to overconfidence, where the overconfidence of STD has also been observed by the existing studies \cite{guo2017calibration,muller2019does}; (3) SD suffers from severe underconfidence; (4) The calibration of CS-KD varies with datasets.
Specifically, it is overconfident on CIFAR-100 but
well calibrated on Tiny ImageNet.

\subsection{Robustness}
As DNNs have been applied to security-critical tasks such as autonomous driving and medical diagnosis, the robustness of DNNs becomes extraordinarily important.
Following the existing study \cite{zhang2016understanding}, we use CIFRA-10 with different levels of corrupted labels to evaluate the robustness of different approaches.
Specifically, we use the open-source implementation \footnote{https://github.com/pluskid/fitting-random-labels} of \cite{zhang2016understanding} to generate four CIFAR-10 training sets, where 20\%, 40\%, 60\%, and 80\% of the labels are replaced with random noise, respectively.
All the test labels are kept intact for evaluation.
%More implementation details including the values of $\alpha$ and $\beta$ are given in the Appendix.
\par
The results are summarized in Table \ref{m5}, where we report the test accuracy in the last epoch (i.e., the 200th epoch) and the best test accuracy achieved during the training process.
First, we notice that there is a large gap between the last accuracy and the best accuracy for most approaches.
The reason is that as the training progresses, the DNN overfits the noise in the later epochs.
\cite{arpit2017closer,han2018co} have a similar observation that DNNs first memorize training data with clean labels and then those with noisy labels.
Especially, when the noise rate is 80\%, the DNN trained with STD overfits a large amount of noise at later epochs, which leads to an extremely large accuracy gap between the last epoch and the best epoch.
However, LWR significantly mitigates this issue as LWR makes use of the information in the past epochs instead of overfitting noise.
As expected, LWR improves the robustness of DNNs significantly.

%Following Zhang et al. (2017), we evaluate the robustness of ERM and mixup models against randomly corrupted labels.
%We hypothesize that increasing the strength of mixup interpolation  should generate virtual examples further from the training examples, making memorization more difficult to achieve. In particular, it should be easier to learn interpolations betoween real examples compared to memorizing interpolations involving random labels. 
%We adpt an open-source implementation (Zhang, 2017)
%to generate four CIFAR-10 training sets, where 20\%, 40\%, 60\% or 80\% of the labels are replaced by random noise, respectively. 
%Following existing studies, We use CIFRA-10 and CIFAR-100 with different levels of corrupted labels to evaluate the robustness of LWR.
%We show that LWR significantly improves the robustness of DNNs to label corruption.

\subsection{Effects of Interval $k$ and Temperature $\tau$}
Interval $k$ denotes updating soft labels once every $k$ epochs, and temperature $\tau$ controls the softness of the labels.
We check their effects on the performance of LWR.
Table \ref{m6} summarizes their effects across different DNN architectures on CIFAR-100.
It is not surprising that when interval $k$ is smaller than a threshold (i.e., the updating frequency is greater than a number), LWR works reasonably well, and when $k$ is set to a large number (e.g., 100), the performances drop significantly.
On the other hand, the best performances of almost all the networks are obtained when $\tau$ is set to 5 rather than 10.
The reason can be that as seen from (\ref{3}), setting $\tau$ to a large number leads to a flat distribution, which may weaken the information in the soft labels.

\section{Conclusion and Future Work}
It is observed that samples in different classes usually share some visual or semantic similarities. However, the typically used one-hot labels cannot capture this kind of information as they are orthogonal for different classes.
On the other hand, we also observe that the standard training procedure discards all the information learned in the past epochs except the weights.
Motivated by these observations, we propose LWR to make use of the learned information in the past to guide the subsequent training.
Specifically, the training logits in the past epochs are used to generate soft labels to provide supervision for the subsequent training.
LWR benefits from label smoothing effects and instance-to-class similarities.
To the end, LWR is able to improve accuracies, calibration, and robustness of DNNs without introducing any network parameters, without any additional inference cost, but only with a negligible training overhead.
Extensive experiments on several datasets have demonstrated the effectiveness of LWR across various modern network architectures.\par

%Different from the standard procedure which discards the past learned information except the weights, we propose to learn with retrospection.
We have applied this idea of learning from the past on classification by proposing LWR.
%by taking advantage of training logits in the past epochs to generate soft labels.
Besides classification, the idea is expected to generalize to regression.
We leave the question on how to use the past learned information in training regression tasks to the future work.

\newpage

\bibliography{lwr.bib}

\begin{thebibliography}{39}
\providecommand{\natexlab}[1]{#1}
\providecommand{\url}[1]{\texttt{#1}}
\providecommand{\urlprefix}{URL }
\expandafter\ifx\csname urlstyle\endcsname\relax
  \providecommand{\doi}[1]{doi:\discretionary{}{}{}#1}\else
  \providecommand{\doi}{doi:\discretionary{}{}{}\begingroup
  \urlstyle{rm}\Url}\fi

\bibitem[{Arpit et~al.(2017)Arpit, Jastrzebski, Ballas, Krueger, Bengio,
  Kanwal, Maharaj, Fischer, Courville, Bengio et~al.}]{arpit2017closer}
Arpit, D.; Jastrzebski, S.~K.; Ballas, N.; Krueger, D.; Bengio, E.; Kanwal,
  M.~S.; Maharaj, T.; Fischer, A.; Courville, A.~C.; Bengio, Y.; et~al. 2017.
\newblock A Closer Look at Memorization in Deep Networks.
\newblock In \emph{ICML}.

\bibitem[{Bagherinezhad et~al.(2018)Bagherinezhad, Horton, Rastegari, and
  Farhadi}]{bagherinezhad2018label}
Bagherinezhad, H.; Horton, M.; Rastegari, M.; and Farhadi, A. 2018.
\newblock Label refinery: Improving imagenet classification through label
  progression.
\newblock \emph{arXiv preprint arXiv:1805.02641} .

\bibitem[{Chorowski and Jaitly(2016)}]{chorowski2016towards}
Chorowski, J.; and Jaitly, N. 2016.
\newblock Towards better decoding and language model integration in sequence to
  sequence models.
\newblock \emph{arXiv preprint arXiv:1612.02695} .

\bibitem[{Dua and Graff(2017)}]{Dua:2019}
Dua, D.; and Graff, C. 2017.
\newblock {UCI} Machine Learning Repository.
\newblock \urlprefix\url{http://archive.ics.uci.edu/ml}.

\bibitem[{Furlanello et~al.(2018)Furlanello, Lipton, Tschannen, Itti, and
  Anandkumar}]{furlanello2018born}
Furlanello, T.; Lipton, Z.~C.; Tschannen, M.; Itti, L.; and Anandkumar, A.
  2018.
\newblock Born again neural networks.
\newblock \emph{arXiv preprint arXiv:1805.04770} .

\bibitem[{Guo et~al.(2017)Guo, Pleiss, Sun, and
  Weinberger}]{guo2017calibration}
Guo, C.; Pleiss, G.; Sun, Y.; and Weinberger, K.~Q. 2017.
\newblock On calibration of modern neural networks.
\newblock \emph{arXiv preprint arXiv:1706.04599} .

\bibitem[{Han et~al.(2018)Han, Yao, Yu, Niu, Xu, Hu, Tsang, and
  Sugiyama}]{han2018co}
Han, B.; Yao, Q.; Yu, X.; Niu, G.; Xu, M.; Hu, W.; Tsang, I.; and Sugiyama, M.
  2018.
\newblock Co-teaching: Robust training of deep neural networks with extremely
  noisy labels.
\newblock In \emph{Advances in neural information processing systems},
  8527--8537.

\bibitem[{He et~al.(2016{\natexlab{a}})He, Zhang, Ren, and Sun}]{he2016deep}
He, K.; Zhang, X.; Ren, S.; and Sun, J. 2016{\natexlab{a}}.
\newblock Deep residual learning for image recognition.
\newblock In \emph{Proceedings of the IEEE conference on computer vision and
  pattern recognition}, 770--778.

\bibitem[{He et~al.(2016{\natexlab{b}})He, Zhang, Ren, and
  Sun}]{he2016identity}
He, K.; Zhang, X.; Ren, S.; and Sun, J. 2016{\natexlab{b}}.
\newblock Identity mappings in deep residual networks.
\newblock In \emph{European conference on computer vision}, 630--645. Springer.

\bibitem[{Hinton, Vinyals, and Dean(2015)}]{hinton2015distilling}
Hinton, G.; Vinyals, O.; and Dean, J. 2015.
\newblock Distilling the knowledge in a neural network.
\newblock \emph{arXiv preprint arXiv:1503.02531} .

\bibitem[{Huang et~al.(2017)Huang, Li, Pleiss, Liu, Hopcroft, and
  Weinberger}]{huang2017snapshot}
Huang, G.; Li, Y.; Pleiss, G.; Liu, Z.; Hopcroft, J.~E.; and Weinberger, K.~Q.
  2017.
\newblock Snapshot ensembles: Train 1, get m for free.
\newblock \emph{arXiv preprint arXiv:1704.00109} .

\bibitem[{Huang et~al.(2019)Huang, Cheng, Bapna, Firat, Chen, Chen, Lee, Ngiam,
  Le, Wu et~al.}]{huang2019gpipe}
Huang, Y.; Cheng, Y.; Bapna, A.; Firat, O.; Chen, D.; Chen, M.; Lee, H.; Ngiam,
  J.; Le, Q.~V.; Wu, Y.; et~al. 2019.
\newblock Gpipe: Efficient training of giant neural networks using pipeline
  parallelism.
\newblock In \emph{Advances in neural information processing systems},
  103--112.

\bibitem[{Khosla et~al.(2011)Khosla, Jayadevaprakash, Yao, and
  Li}]{khosla2011novel}
Khosla, A.; Jayadevaprakash, N.; Yao, B.; and Li, F.-F. 2011.
\newblock Novel dataset for fine-grained image categorization: Stanford dogs.
\newblock In \emph{Proc. CVPR Workshop on Fine-Grained Visual Categorization
  (FGVC)}, volume~2.

\bibitem[{Kingma and Ba(2015)}]{kingma2014adam}
Kingma, D.~P.; and Ba, J. 2015.
\newblock Adam: A method for stochastic optimization.
\newblock \emph{International Conference for Learning Representations} .

\bibitem[{Krizhevsky and Hinton(2009)}]{krizhevsky2009learning}
Krizhevsky, A.; and Hinton, G. 2009.
\newblock Learning multiple layers of features from tiny images.
\newblock Technical report, Citeseer.

\bibitem[{Lawrence et~al.(1997)Lawrence, Giles, Tsoi, and
  Back}]{lawrence1997face}
Lawrence, S.; Giles, C.~L.; Tsoi, A.~C.; and Back, A.~D. 1997.
\newblock Face recognition: A convolutional neural-network approach.
\newblock \emph{IEEE transactions on neural networks} 8(1): 98--113.

\bibitem[{Levinson et~al.(2011)Levinson, Askeland, Becker, Dolson, Held,
  Kammel, Kolter, Langer, Pink, Pratt et~al.}]{levinson2011towards}
Levinson, J.; Askeland, J.; Becker, J.; Dolson, J.; Held, D.; Kammel, S.;
  Kolter, J.~Z.; Langer, D.; Pink, O.; Pratt, V.; et~al. 2011.
\newblock Towards fully autonomous driving: Systems and algorithms.
\newblock In \emph{2011 IEEE Intelligent Vehicles Symposium (IV)}, 163--168.
  IEEE.

\bibitem[{Loshchilov and Hutter(2017)}]{loshchilov2016sgdr}
Loshchilov, I.; and Hutter, F. 2017.
\newblock Sgdr: Stochastic gradient descent with warm restarts.
\newblock \emph{International Conference for Learning Representations} .

\bibitem[{Ma et~al.(2018)Ma, Zhang, Zheng, and Sun}]{ma2018shufflenet}
Ma, N.; Zhang, X.; Zheng, H.-T.; and Sun, J. 2018.
\newblock Shufflenet v2: Practical guidelines for efficient cnn architecture
  design.
\newblock In \emph{Proceedings of the European conference on computer vision
  (ECCV)}, 116--131.

\bibitem[{Maji et~al.(2013)Maji, Kannala, Rahtu, Blaschko, and
  Vedaldi}]{maji13fine-grained}
Maji, S.; Kannala, J.; Rahtu, E.; Blaschko, M.; and Vedaldi, A. 2013.
\newblock Fine-Grained Visual Classification of Aircraft.
\newblock Technical report.

\bibitem[{Miotto et~al.(2016)Miotto, Li, Kidd, and Dudley}]{miotto2016deep}
Miotto, R.; Li, L.; Kidd, B.~A.; and Dudley, J.~T. 2016.
\newblock Deep patient: an unsupervised representation to predict the future of
  patients from the electronic health records.
\newblock \emph{Scientific reports} 6(1): 1--10.

\bibitem[{M{\"u}ller, Kornblith, and Hinton(2019)}]{muller2019does}
M{\"u}ller, R.; Kornblith, S.; and Hinton, G.~E. 2019.
\newblock When does label smoothing help?
\newblock In \emph{Advances in Neural Information Processing Systems},
  4694--4703.

\bibitem[{Pereyra et~al.(2017)Pereyra, Tucker, Chorowski, Kaiser, and
  Hinton}]{pereyra2017regularizing}
Pereyra, G.; Tucker, G.; Chorowski, J.; Kaiser, {\L}.; and Hinton, G. 2017.
\newblock Regularizing neural networks by penalizing confident output
  distributions.
\newblock \emph{arXiv preprint arXiv:1701.06548} .

\bibitem[{Real et~al.(2019)Real, Aggarwal, Huang, and Le}]{real2019regularized}
Real, E.; Aggarwal, A.; Huang, Y.; and Le, Q.~V. 2019.
\newblock Regularized evolution for image classifier architecture search.
\newblock In \emph{Proceedings of the aaai conference on artificial
  intelligence}, volume~33, 4780--4789.

\bibitem[{Sandler et~al.(2018)Sandler, Howard, Zhu, Zhmoginov, and
  Chen}]{sandler2018mobilenetv2}
Sandler, M.; Howard, A.; Zhu, M.; Zhmoginov, A.; and Chen, L.-C. 2018.
\newblock Mobilenetv2: Inverted residuals and linear bottlenecks.
\newblock In \emph{Proceedings of the IEEE conference on computer vision and
  pattern recognition}, 4510--4520.

\bibitem[{Simonyan and Zisserman(2014)}]{simonyan2014very}
Simonyan, K.; and Zisserman, A. 2014.
\newblock Very deep convolutional networks for large-scale image recognition.
\newblock \emph{arXiv preprint arXiv:1409.1556} .

\bibitem[{Szegedy et~al.(2016)Szegedy, Vanhoucke, Ioffe, Shlens, and
  Wojna}]{szegedy2016rethinking}
Szegedy, C.; Vanhoucke, V.; Ioffe, S.; Shlens, J.; and Wojna, Z. 2016.
\newblock Rethinking the inception architecture for computer vision.
\newblock In \emph{Proceedings of the IEEE conference on computer vision and
  pattern recognition}, 2818--2826.

\bibitem[{Vaswani et~al.(2017)Vaswani, Shazeer, Parmar, Uszkoreit, Jones,
  Gomez, Kaiser, and Polosukhin}]{vaswani2017attention}
Vaswani, A.; Shazeer, N.; Parmar, N.; Uszkoreit, J.; Jones, L.; Gomez, A.~N.;
  Kaiser, {\L}.; and Polosukhin, I. 2017.
\newblock Attention is all you need.
\newblock In \emph{Advances in neural information processing systems},
  5998--6008.

\bibitem[{Wah et~al.(2011)Wah, Branson, Welinder, Perona, and
  Belongie}]{wah2011caltech}
Wah, C.; Branson, S.; Welinder, P.; Perona, P.; and Belongie, S. 2011.
\newblock The caltech-ucsd birds-200-2011 dataset .

\bibitem[{Xu and Liu(2019)}]{xu2019data}
Xu, T.-B.; and Liu, C.-L. 2019.
\newblock Data-distortion guided self-distillation for deep neural networks.
\newblock In \emph{Proceedings of the AAAI Conference on Artificial
  Intelligence}, volume~33, 5565--5572.

\bibitem[{Yang et~al.(2019{\natexlab{a}})Yang, Xie, Qiao, and
  Yuille}]{yang2019training}
Yang, C.; Xie, L.; Qiao, S.; and Yuille, A.~L. 2019{\natexlab{a}}.
\newblock Training deep neural networks in generations: A more tolerant teacher
  educates better students.
\newblock In \emph{Proceedings of the AAAI Conference on Artificial
  Intelligence}, volume~33, 5628--5635.

\bibitem[{Yang et~al.(2019{\natexlab{b}})Yang, Xie, Su, and
  Yuille}]{yang2019snapshot}
Yang, C.; Xie, L.; Su, C.; and Yuille, A.~L. 2019{\natexlab{b}}.
\newblock Snapshot distillation: Teacher-student optimization in one
  generation.
\newblock In \emph{Proceedings of the IEEE Conference on Computer Vision and
  Pattern Recognition}, 2859--2868.

\bibitem[{Yuan et~al.(2020)Yuan, Tay, Li, Wang, and Feng}]{yuan2020revisiting}
Yuan, L.; Tay, F.~E.; Li, G.; Wang, T.; and Feng, J. 2020.
\newblock Revisiting Knowledge Distillation via Label Smoothing Regularization.
\newblock In \emph{Proceedings of the IEEE/CVF Conference on Computer Vision
  and Pattern Recognition}, 3903--3911.

\bibitem[{Yun et~al.(2020)Yun, Park, Lee, and Shin}]{yun2020regularizing}
Yun, S.; Park, J.; Lee, K.; and Shin, J. 2020.
\newblock Regularizing class-wise predictions via self-knowledge distillation.
\newblock In \emph{Proceedings of the IEEE/CVF Conference on Computer Vision
  and Pattern Recognition}, 13876--13885.

\bibitem[{Zagoruyko and Komodakis(2016)}]{Zagoruyko2016WRN}
Zagoruyko, S.; and Komodakis, N. 2016.
\newblock Wide Residual Networks.
\newblock In \emph{BMVC}.

\bibitem[{Zhang et~al.(2017)Zhang, Bengio, Hardt, Recht, and
  Vinyals}]{zhang2016understanding}
Zhang, C.; Bengio, S.; Hardt, M.; Recht, B.; and Vinyals, O. 2017.
\newblock Understanding deep learning requires rethinking generalization.
\newblock \emph{International Conference for Learning Representations} .

\bibitem[{Zhang et~al.(2018)Zhang, Cisse, Dauphin, and
  Lopez-Paz}]{zhang2017mixup}
Zhang, H.; Cisse, M.; Dauphin, Y.~N.; and Lopez-Paz, D. 2018.
\newblock mixup: Beyond empirical risk minimization.
\newblock \emph{International Conference for Learning Representations} .

\bibitem[{Zhang et~al.(2019)Zhang, Song, Gao, Chen, Bao, and
  Ma}]{zhang2019your}
Zhang, L.; Song, J.; Gao, A.; Chen, J.; Bao, C.; and Ma, K. 2019.
\newblock Be your own teacher: Improve the performance of convolutional neural
  networks via self distillation.
\newblock In \emph{Proceedings of the IEEE International Conference on Computer
  Vision}, 3713--3722.

\bibitem[{Zoph et~al.(2018)Zoph, Vasudevan, Shlens, and Le}]{zoph2018learning}
Zoph, B.; Vasudevan, V.; Shlens, J.; and Le, Q.~V. 2018.
\newblock Learning transferable architectures for scalable image recognition.
\newblock In \emph{Proceedings of the IEEE conference on computer vision and
  pattern recognition}, 8697--8710.

\end{thebibliography}

\end{document}